\documentclass{article}

\usepackage{arxiv}

\usepackage{xcolor}
\usepackage{epsfig}
\usepackage{subcaption}
\usepackage{calc}
\usepackage{amssymb}
\usepackage{amstext}
\usepackage{amsmath}
\usepackage{amsthm}
\usepackage{multicol}
\usepackage{pslatex}
\usepackage{apalike}
\usepackage{algorithm2e}

\usepackage[utf8]{inputenc} 
\usepackage[T1]{fontenc}    
\usepackage{hyperref}       
\usepackage{url}            
\usepackage{booktabs}       
\usepackage{amsfonts}       
\usepackage{nicefrac}       
\usepackage{microtype}      
\usepackage{lipsum}
\usepackage{graphicx}

\graphicspath{ {./images/} }

\title{Enterprise Resource Planning Using Multi-type Transformers in Ferro-Titanium Industry}

\author{
 Samira Yazdanpourmoghadam \\
  Department of Mathematics and Industrial Engineering\\
  Polytechnique Montreal\\
  Montreal, Canada \\
  \texttt{samira.yazdanpourmoghadam@polymtl.ca} \\
   \And
 Mahan Balal Pour \\
  Department of Mechanical, Industrial and Aerospace Engineering\\
  Concordia University\\
  Montreal, Canada \\
  \texttt{m\textunderscore balalp@live.concordia.ca} \\
  \And
 Vahid Partovi Nia \\
  Department of Mathematics and Industrial Engineering\\
  Polytechnique Montreal\\
  Montreal, Canada \\
  \texttt{vahid.partovinia@polymtl.ca} \\
}

\begin{document}
\maketitle
\begin{abstract}
Combinatorial optimization problems such as the Job-Shop Scheduling Problem (JSP) and Knapsack Problem (KP) are fundamental challenges in operations research,  logistics, and eterprise resource planning (ERP). These problems often require sophisticated algorithms to achieve near-optimal solutions within practical time constraints. Recent advances in deep learning have introduced transformer-based architectures as promising alternatives to traditional heuristics and metaheuristics. We leverage the Multi-Type Transformer (MTT) architecture to address these benchmarks in a unified framework.  We present an extensive experimental evaluation across standard benchmark datasets for JSP and KP, demonstrating that MTT achieves competitive performance on different size of these benchmark problems. We showcase the potential of multi-type attention on a real application in Ferro-Titanium industry. To the best of our knowledge, we are the first to apply multi-type transformers in real manufacturing.\end{abstract}


\section{\uppercase{Introduction}}
\label{sec:introduction}

Combinatorial optimization lies at the core of numerous real-world applications, including manufacturing scheduling, transportation planning, resource allocation,  agentic artificial intelligence, and robotics. Problems such as the Job-Shop Scheduling Problem (JSP) and Knapsack Problem (KP) are widely studied due to their computational complexity and practical relevance. Traditional approaches—ranging from exact algorithms to heuristic and metaheuristic methods often struggle to balance solution quality and computational efficiency, particularly for large-scale instances.
The emergence of neural combinatorial optimization has introduced a paradigm shift, leveraging deep learning models to learn problem structures and generate high-quality solutions. Transformer-based architectures, in particular, have demonstrated remarkable success in sequence modelling and graph-based representations, making them suitable for optimization tasks. We adopt the Multi-Type Transformer (MTT) architecture introduced in \cite{Drakulic_GOAL_ICLR2025}. MTT extends the conventional transformer by integrating multiple attention types, each specialized for distinct structural aspects of the input, thereby enhancing representational capacity and adaptability. We investigate the effectiveness of MTT in solving different sizes of JSP and KP within a unified framework to study the generalizability and scaling of this new tool.

Neural combinatorial optimization has emerged as a promising alternative to classical heuristics for solving NP-hard problems such as JSP, and KP. Early works like  \cite{Kool_AttentionRouting_ICLR2018} introduced attention-based models and reinforcement learning for routing and scheduling tasks. These approaches demonstrated that deep learning can learn problem structures and generate competitive solutions without handcrafted heuristics. Transformers are developed for natural language processing, and have been adapted for combinatorial optimization due to their ability to model long-range dependencies. Standard transformer models employ a single attention mechanism, which works well for homogeneous input structures but struggles with heterogeneous entities common in combinatorial optimization, e.g. jobs and machines in JSP.

\begin{figure}[t]
    \centering
    \includegraphics[width=\linewidth]{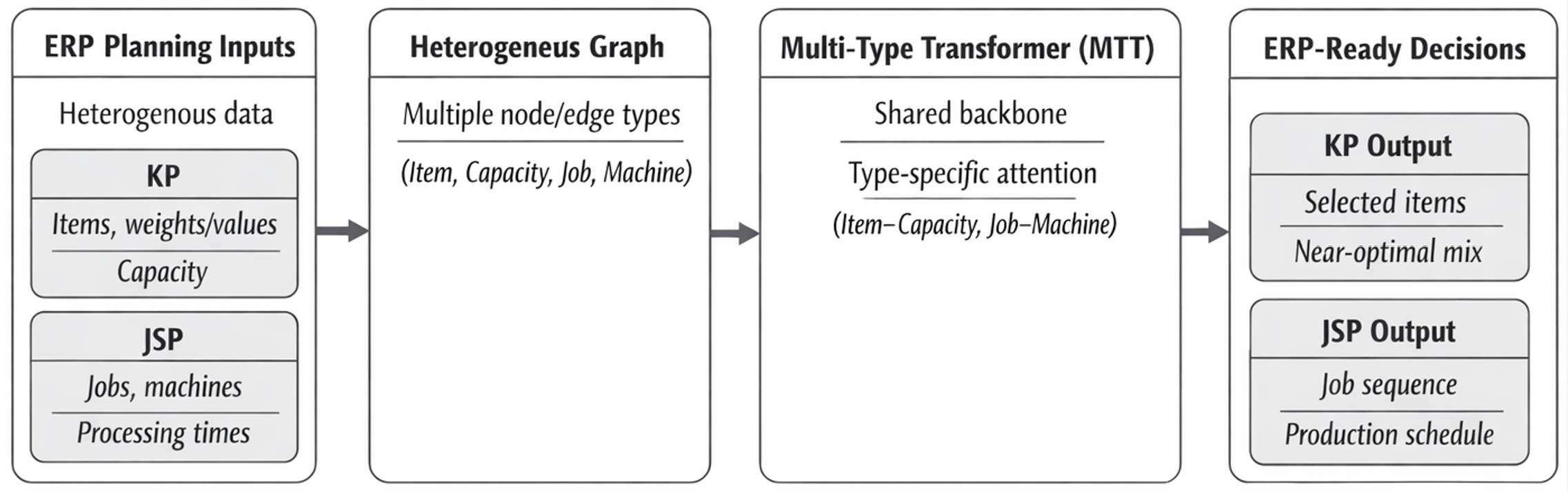}
    \caption{Overview of the unified MTT pipeline for ERP optimization, KP and JSP are represented using heterogeneous graphs and solved using a shared Transformer backbone with type-specific attention.}
    \label{fig:multitype}
\end{figure}

The advent of neural combinatorial optimization reframed discrete decision-making as sequence modelling and structured prediction. Pointer Networks introduced attention-based pointing to variable-length outputs, enabling end-to-end learning for geometric and routing problems \cite{Vinyals2015}. Transformers abstracted this idea to a fully attention-based architecture, dispensing with recurrence and convolution while improving scalability and parallelism \cite{Vaswani2017}. Reinforcement learning approaches subsequently specialized to routing with dynamic state representations, as in \cite{Nazari2018}. \cite{Bello2017} established early evidence that policy gradients can discover competitive heuristics without handcrafted rules.

Graph representation learning further advanced neural solvers by encoding the structure of discrete problems directly into the model. \cite{Dai2017} combined reinforcement learning with graph embeddings \emph{Structure2Vec} to learn greedy meta-algorithms for Minimum Vertex Cover, and Max-Cut. Complementary lines applied graph neural networks and specialized architectures to routing and packing tasks, including efficient encoders and problem-aware attention mechanisms \cite{Joshi2019}. Together, these approaches improved generalization across instance distributions, revealing inductive biases that are critical for scalability and transfer.

More recently, generalist models aim to unify heterogeneous optimization tasks within a single backbone. Multi-type transformer designs and shared representations allow a solver to tackle knapsack, routing, and scheduling with minimal task-specific customization. Examples include the MTT framework and unified architectures that pursue cross-problem learning and lightweight adapters for rapid adaptation. In ERP settings, such generalist solvers are attractive because they can reuse learned abstractions across procurement, blending, routing, and shop-floor scheduling while maintaining acceptable optimality gaps relative to exact baselines \cite{Kellerer2004,Applegate2007,Pinedo2012}.

The Multi-Type Transformer (MTT) introduced in GOAL \cite{Drakulic_GOAL_ICLR2025} addresses this limitation by incorporating multiple attention types within the same architecture, see Figure~\ref{fig:multitype}. Each attention type specializes in capturing relationships among distinct entity types, such as job-machine pairs in JSP or item-capacity pairs in KP. This design improves representation learning and enables generalization across diverse combinatorial optimization problems. Recent works like \cite{Zong_UniCO_2025} have demonstrated that multi-type attention can outperform single-type models in both accuracy and adaptability. GOAL integrates multi-type attention blocks with parameter sharing, achieving strong performance across routing, scheduling, and packing tasks without retraining from scratch. 

MTT offer heterogeneous representation learning for complex problem structures, offers cross-domain generalization, and reduces problem-specific customization. However, challenges remain in computational efficiency for large-scale instances and integration with reinforcement learning for dynamic decision-making.

\section{\uppercase{Methodology}}
We solve two canonical problems, KP and JSP, within a unified graph formulation that exposes heterogeneous node-edge types. For JSP, we use a disjunctive graph with operation nodes, conjunctive edges (technological precedence), and disjunctive edges per machine capturing mutual exclusivity; features include processing times, machine IDs, release/ready times, and partial schedules (dynamic state). This follows established JSP graph formulations in NCO and RL scheduling research, adapted for transformer encoders to avoid hand‑engineered features.


For KP, we formulate a bipartite graph between items and a single capacity node, where item nodes carry weight and value, and a capacity node accumulates residual capacity; a feasibility mask prevents selections violating capacity. This packing representation is consistent with generalist NCO abstractions that unify heterogeneous COPs.  This multi‑relational graph makes heterogeneity explicit: jobs versus machines (JSP), and items versus capacity (KP), enabling type‑specific attention.


\begin{figure}
    \centering
    \fbox{\includegraphics[height=0.2\textheight]{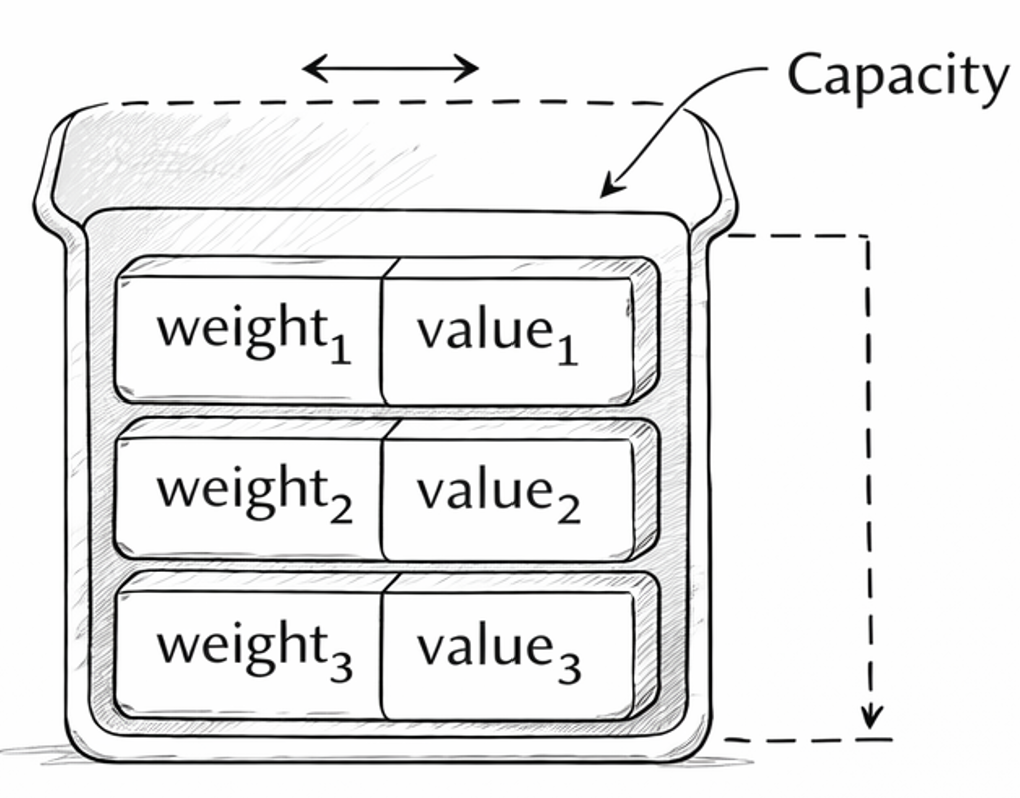}}    \fbox{\includegraphics[height=0.2\textheight]{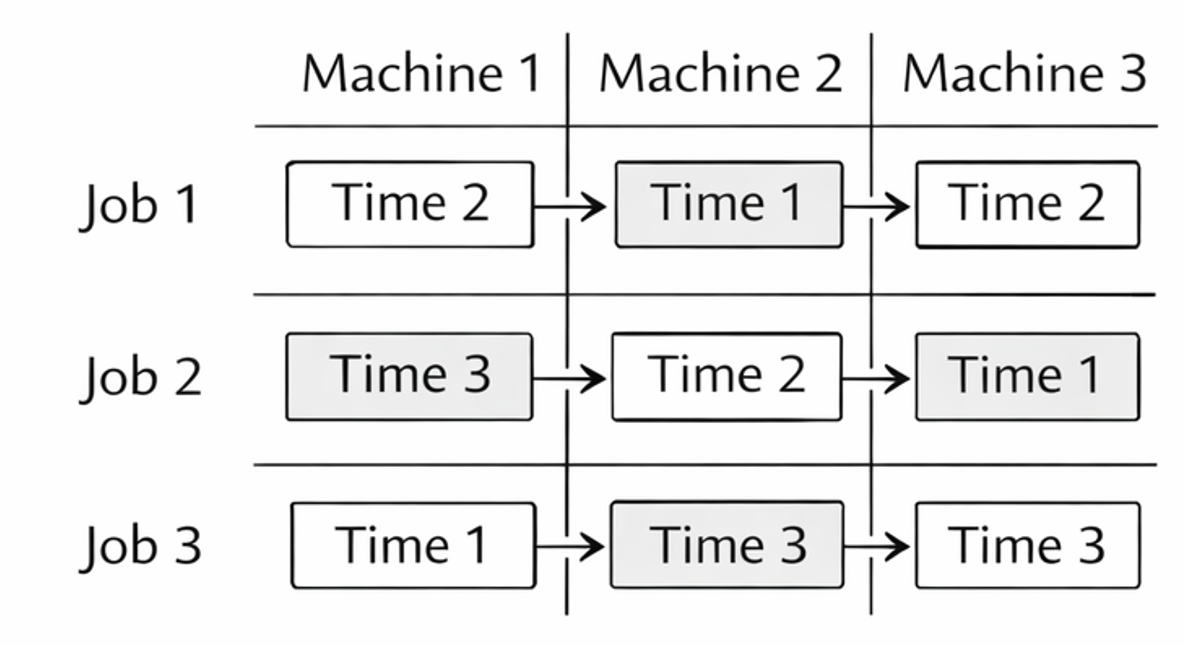}}
    \caption{Schematic illustration of combinatorial problems studied, the visualization of KP (left panel), and JSP (right panel).}
    \label{fig:kpjsp}
\end{figure}

\section{\uppercase{Combinatorial Benchmarks}}
\label{sec:combinatorial}
This section introduces two canonical NP-hard combinatorial optimization problems that frequently arise in ERP and decision-making systems: the (0--1) Knapsack and the Job-Shop Scheduling Problem (JSP). We present their standard mathematical programming formulations, simple visual examples, and discuss their challenges, similarities, and differences for readers familiar with undergraduate-level mathematical programming.

\subsection{Knapsack (KP)}
KP is about having a limited capacity (like weight, budget, time, battery, memory), and a list of things one could include. Each item gives some benefit (usefulness, satisfaction, profit). The problem is about picking the best combination that fits within the limit and gets the most benefit overall. Here we focus on all‑or‑nothing (0–1 knapsack) whether or not bring the whole item, see Figure~\ref{fig:kpjsp} (left panel).
Given $n$ items with values $v_i>0$ and weights $w_i>0$, and a capacity $W>0$, choose a subset that maximizes total value without exceeding the capacity of the sack $W$. Let $x_i \in \{0,1\}$ indicate whether item $i$ is selected i.e. a binary variable $x_i\in\{0,1\}$.

\begin{align}
\max_{x \in \{0,1\}^n} \quad & \sum_{i=1}^n v_i x_i \nonumber\\
\text{s.t.}\quad & \sum_{i=1}^n w_i x_i \le W.\nonumber 
\end{align}
Think of a backpack of capacity $W$ and items as bars with lengths $w_i$ (weight) and heights $v_i$ (value). The aim is to pack bars so their total length does not exceed $W$, while their total height (sum of values) is maximized. As a simple instance, suppose the backpack of $W=7$ capacity and four items:
\[
(w,v) = \{(2,3),~(3,4),~(4,5),~(5,8)\}.
\]
Feasible selections include:
\begin{itemize}\itemsep0.2em
\item $x=[1,0,0,1]$: total weight $2+5=7$ (fits in the sack of capacity 8, with total value $3+8=11$ so is a feasible solution.
\item $x=[0,1,1,0]$: weight $3+4=7$, value $4+5=9$ is also a feasible but with a lower total value.
\item $x=[1,1,1,0]$: weight $9$ which is infeasible because it exceeds the total capacity $W=8$.
\end{itemize}
The best among these is $x=[1,0,0,1]$ with value $11$, which is optimal for this instance.

KP is NP-hard; however, it admits pseudo-polynomial dynamic programming (DP) in $O(nW)$ and a fully polynomial-time approximation scheme. Integer programming solvers (branch-and-bound/cut), Lagrangian relaxation, and learned heuristics can efficiently solve large instances arising in ERP.

\subsection{Job-Shop Scheduling Problem (JSP)}
You have multiple jobs (customer orders or work orders).
Each job must go through a specific sequence of steps (operations) like \texttt{paint -> assemble -> test}. Each step must run on a specific machine/work center, which can only do one thing at a time.
You want to pick a good sequence and start times so that key goals are met (e.g., deliver on time, finish sooner, reduce waiting/WIP, keep machines utilized). The problem is about the order we should run tasks on shared machines such that every job finishes efficiently and on time, see Figure~\ref{fig:kpjsp} (right panel). JSP has several concrete applications in ERP. In resource planning 
\begin{itemize}
    \item Jobs are customer orders or production orders.
    \item Operations are steps like cutting, milling, painting, assembly, test
Machines/work.
\item centers are resources in ERP PP/Shop Floor Control
Processing times.
\item  setups are duration and changeover overhead.
\item Constraints are precedence (step order), calendars/shifts, machine downtime
\item Objectives are on‑time delivery (tardiness), total completion time, utilization, sequence stability.
\end{itemize}

Suppose there are $J$ jobs, each consisting of an ordered sequence of operations that must be processed on specific machines from the set $\mathcal{M}=\{1,\dots,M\}$. Operation $o$ has processing time $p_o>0$ and is assigned to a machine $m(o) \in \mathcal{M}$. Each machine processes at most one operation at a time. The goal is to minimize the makespan (completion time of the last job). Let's start with a simple example 
Jobs operation to machine (time):
\begin{itemize}
    \item  J1: Cut → M1 (3h) → Paint → M2 (2h) → Test → M3 (2h)
\item J2: Prep → M2 (2h) → Inspect → M3 (1h) → Assemble → M1 (4h)
\item J3: Forge → M3 (3h) → Drill → M1 (2h) → Coat → M2 (3h)
\end{itemize}
A feasible schedule might look like this:

Machine M1 (one job at a time)
\begin{itemize}
    \item [0–3] J1-Cut   → 3h
\item [3–5] J3-Drill → 2h
\item [5–9] J2-Assem → 4h
\end{itemize}

Machine M2 (one job at a time)
\begin{itemize}
\item [0–2] J2-Prep  → 2h
\item [3–5] J1-Paint → 2h   (waited until J1-Cut finished at t=3)
\item [5–8] J3-Coat  → 3h
\end{itemize}

Machine M2 (one job at a time)
\begin{itemize}
\item [0–3] J3-Forge → 3h
\item [3–4] J2-Inspect → 1h (after J2-Prep finished at t=2)
\item [5–7] J1-Test  → 2h   (after J1-Paint finished at t=5)
\end{itemize}
Conflicting priorities: Many jobs want the same machine next. You must choose who goes first.
Precedence chains: You can’t start “paint” before “cut”—delays ripple through the chain.
Setup/changeovers: Switching colours/tools adds overhead; the ``best'' sequence may minimize changeovers.
Calendars and disruptions: Shifts, breaks, maintenance, rush orders, and no‑shows complicate plans.
Combinatorial explosion: Trying all possible sequences is infeasible—even for modest shop sizes.

However, a better quality solution can be provided using heuristics and metaheuristics methods such as
Shifting bottleneck, tabu search, simulated annealing, genetic algorithms—great for complex shops.
Below, we explain the standard Optimization and constraint programming, which is used here to benchmark our transformer-based solution.  This constrained programming is finite capacity scheduling with disjunctive constraints, time‑indexed models, and constrained programming to handle precedence and calendars.

Learning‑based scheduling (data‑driven) is what our transformer-based model is trying to do.  Predict good priorities or sequence patterns (e.g., using historical data, graph neural networks over job‑machine networks), then feed to ERP as priority rules or warm starts.

JSP is strongly NP-hard; the search space grows factorially with machine sequences. Common methods include IP/CP formulations, branch-and-bound with strong cuts, time-indexed models, disjunctive-graph/critical-path local search, metaheuristics (tabu, SA, GA), and learning to schedule (e.g., learned dispatching rules or graph neural networks over the disjunctive graph). For pattern recognition workflows, JSSP naturally models the sequencing of compute stages on shared accelerators or pipelines with precedence.

Let $S_o \ge 0$ be the start time of operation $o$ and $C_{\max}$ the makespan. For any two operations $a,b$ that require the same machine $m$, introduce a binary ordering variable $y_{ab}\in\{0,1\}$ (``$a$ before $b$'' vs. ``$b$ before $a$''). Let $\mathcal{P}(j)$ denote the ordered operations of job $j$ (predecessor $\mathrm{pred}(o)$ if it exists). With a sufficiently large constant $M$:
\begin{align}
\min \quad & C_{\max} \label{eq:jssp_obj} \\\
\text{s.t.}\quad
& S_o \ge S_{\mathrm{pred}(o)} + p_{\mathrm{pred}(o)} \quad \forall o \text{ with predecessor} \label{eq:jssp_prec}\\
& S_a + p_a \le S_b + M(1-y_{ab}) \quad \forall (a,b): m(a)=m(b) \label{eq:jssp_machine1}\\
& S_b + p_b \le S_a + M(y_{ab}) \quad \forall (a,b): m(a)=m(b) \label{eq:jssp_machine2}\\
& C_{\max} \ge S_o + p_o \quad \forall o \label{eq:jssp_cmax}\\
& S_o \ge 0,\quad y_{ab}\in\{0,1\}. \nonumber
\end{align}
Constraints \eqref{eq:jssp_prec} enforce job order; \eqref{eq:jssp_machine1}--\eqref{eq:jssp_machine2} enforce non-overlap on machines by choosing one of two possible orderings; \eqref{eq:jssp_cmax} defines the makespan.

\paragraph{Illustrative example (3 jobs, 3 machines).}
Jobs and routes:
\[
\begin{aligned}
J_1 &: (M_1,3)\rightarrow(M_2,2)\rightarrow(M_3,2)\\
J_2 &: (M_2,2)\rightarrow(M_3,1)\rightarrow(M_1,4)\\
J_3 &: (M_3,3)\rightarrow(M_1,2)\rightarrow(M_2,3)
\end{aligned}
\]
One feasible schedule (makespan $C_{\max}=9$) is:
\begin{center}\tiny
\begin{tabular}{l}
\texttt{Machine M1: [0--3] J1-1, [3--5] J3-2, [5--9] J2-3}\\
\texttt{Machine M2: [0--2] J2-1, [3--5] J1-2, [5--8] J3-3}\\
\texttt{Machine M3: [0--3] J3-1, [3--4] J2-2, [5--7] J1-3}
\end{tabular}
\end{center}
Each job respects its precedence; machines never process more than one operation at a time.

\subsection{Problem Similarities}
Both KP and JSP problems face enormous combinatorial search spaces and tight feasibility constraints. KP's real cases often include multi-dimensional capacities, conflicts, or groups, which destroy simple structure. JSP's temporal precedence coupling and machine disjunctions loosen linear relaxations. In data-driven contexts, instance distributions may drift, requiring learned heuristics to generalize and remain robust to noise and uncertainty.

In terms of the similarities between the two problems, both KP and JSP  (i) Binary and ordinal decisions drive feasibility (select-or-not; before-or-after). (ii) Objectives are additive but constrained: knapsack by capacity, JSP by non-overlap and precedence. (iii) Both benefit from branch-and-bound, cutting planes, and surrogate learning.

In terms of differences, KP is NP-hard with pseudo-polynomial Dynamic Programming and a Fully Polynomial‑Time Approximation Scheme, while JSP is strongly NP-hard without such a Dynamic Programming structure. KP decisions are static subsets, while JSP decisions are dynamic sequences with time. Relaxations for KP, e.g., fractional Linear Programming, are tighter and yield good bounds, while JSP relaxations often require sophisticated cuts or decomposition, such as disjunctive graphs, time-indexed formulations. Visually, KP is capacity packing, and  JSP is a temporal machine assignment with a Gantt-like structure.

Feature representations of instances, such as item attributes, job-machine graphs, enable learned heuristics that predict promising orderings or selections, prune search, and provide strong warm starts for exact solvers. Embedding disjunctive graphs via Transformers, Graph Neural Networks, or learning value-to-weight surrogates for knapsack, can reduce solver time substantially while maintaining solution quality, especially when training on synthetic or historical instance distributions.

While we present canonical forms, many variants  such as multi-dimensional, bi-criteria knapsack, flexible, job-blocking JSSP arise. The formulations above are easily adapted by adding constraints or variables to capture such specifics.





\section{\uppercase{Benchmarking}}
Running combinatorial optimization problems such as the KP and JSP requires scalable algorithms to handle varying instance complexities. In our experiments, the KP is evaluated on item sets of sizes 50, 60, 70, 80, 90, and 100, reflecting increasing dimensionality and computational demands for resource allocation under capacity constraints. These choices reflect scenarios that we encounter in the Ferro-Titanium industry. 
We explain the KP more in detail as our application in the Ferro-Titanium industry is focused on this case, but we also evaluated the JSP too.
\begin{figure}[t]
    \centering
    \includegraphics[width=0.45\linewidth]{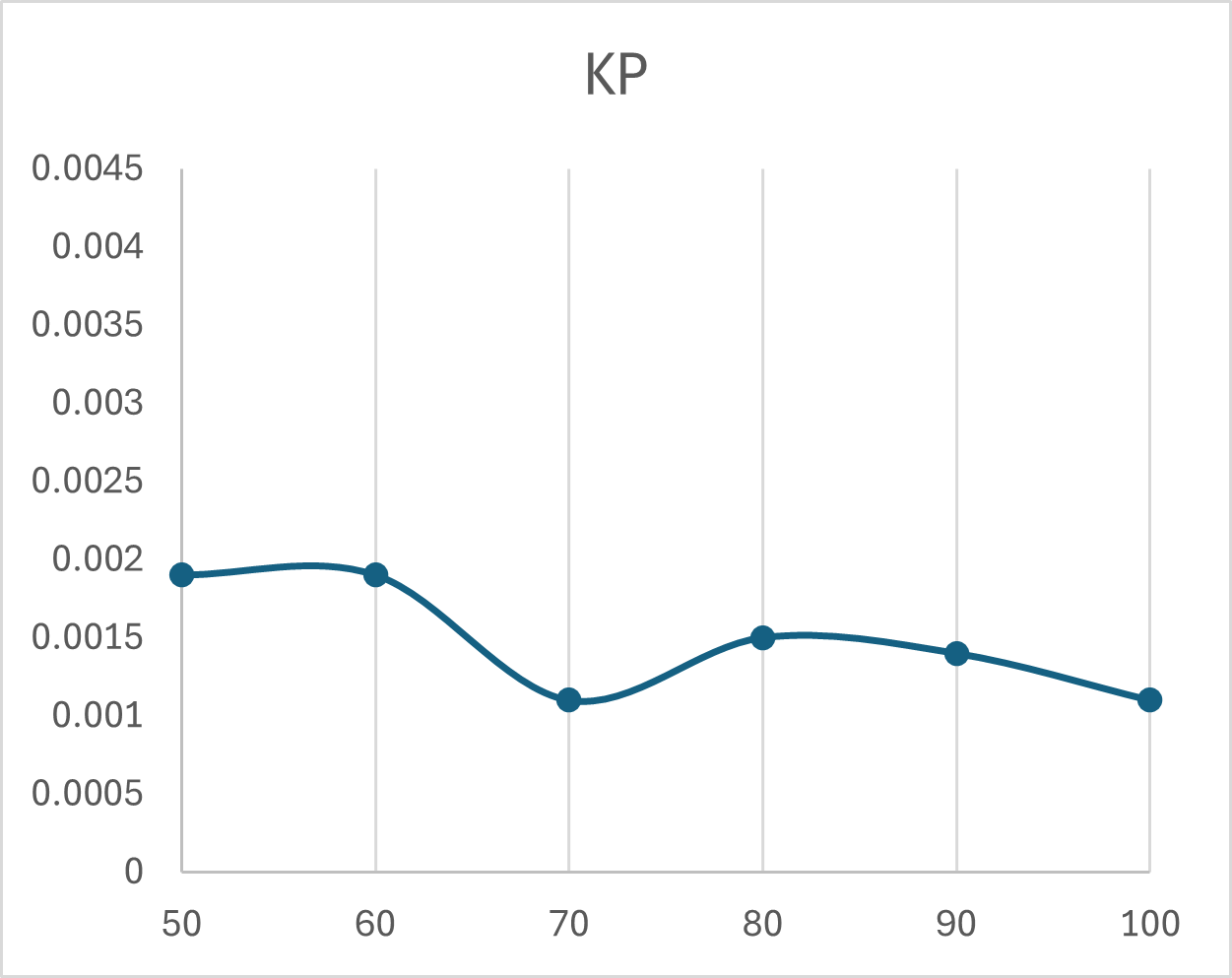}    \includegraphics[width=0.45\linewidth]{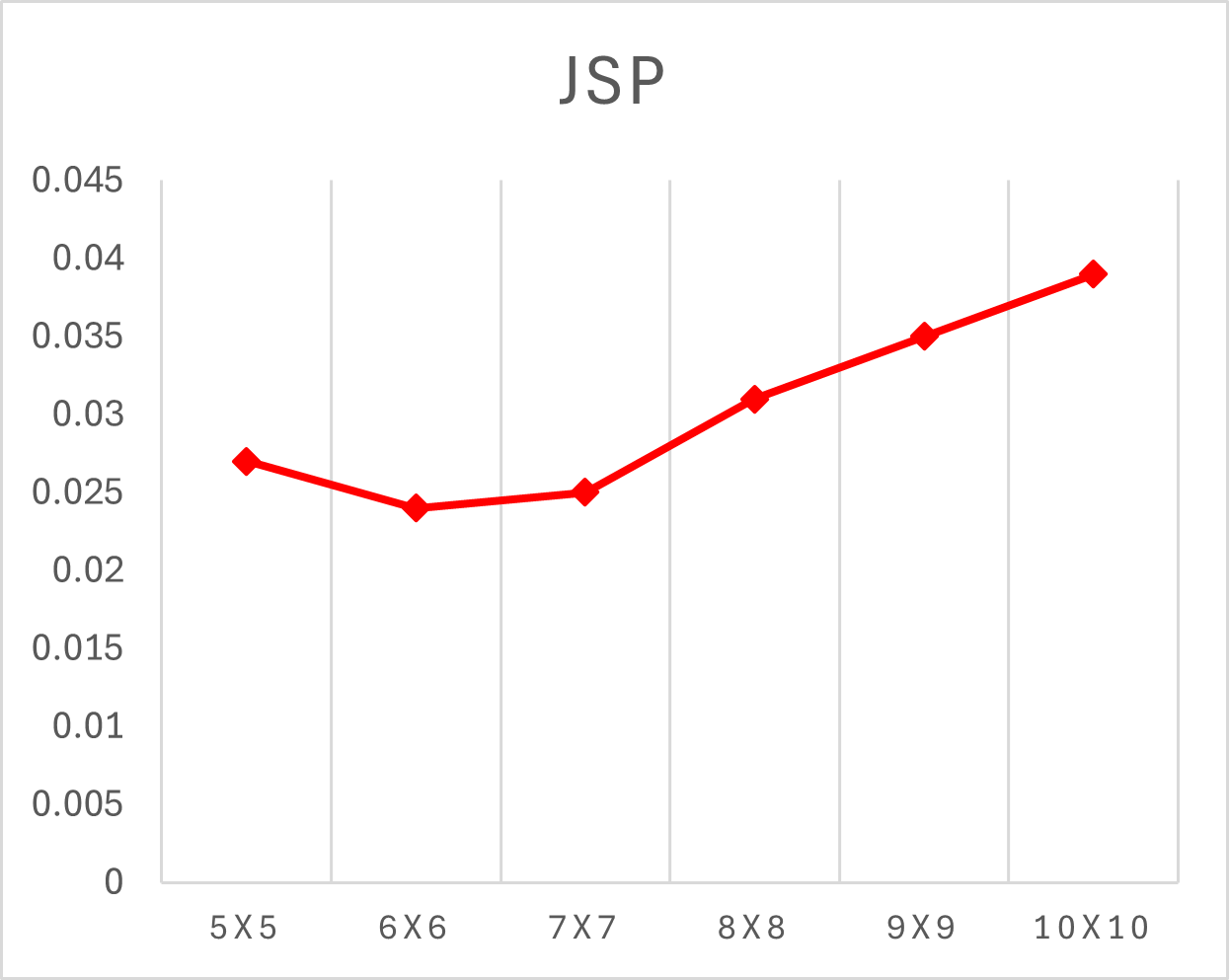}
    \caption{The optimality gap of the OR solver compared with MTT on KP and JSP problems, see Table~\ref{tab:bench}.}
    \label{fig:placeholder}
\end{figure}

Combinatorial optimization addresses decision-making problems where the objective is to find an optimal solution from a finite but exponentially large set of possibilities, subject to specific constraints. These problems arise in diverse domains such as logistics, manufacturing, and resource allocation, and are often NP-hard, requiring sophisticated algorithms to achieve near-optimal solutions efficiently. Two canonical examples are the Knapsack Problem, which models resource allocation under capacity constraints, and the Job-Shop Scheduling Problem, which focuses on sequencing tasks across machines to minimize completion time. Both exemplify the complexity and practical relevance of combinatorial optimization, motivating the development of advanced methods such as metaheuristics and learning-based approaches to tackle real-world challenges. Here we apply the problem to the Knapsack Ferro-titatium industry that we faced in the industry.

The 0/1 Knapsack Problem is categorized as a Packing problem for a single Sack. It is defined as a combinatorial optimization problem where the agent must select a subset of items from a given set. 
\begin{itemize}
    \item Constraints: Each item has a specific weight ($w_i$) and value ($v_i$). The selection must not exceed a predefined cumulative weight threshold, known as the ``knapsack capacity", maximum capacity (C). 
    \item Objective: The goal is to maximize the total accumulated value of the selected items without breaking the capacity constraint.
In the test set, we utilized a statistically significant subset of 128 instances.
\end{itemize}

This sample size is consistent with the validation methodology used in the GOAL paper to ensure statistical significance when reporting the Optimality Gap. For our specific experiment, custom datasets including N=100, 90, 80, 70 items were generated, testing the model's ability to generalize to unseen problem sizes. Data Structure and  Formatting
To ensure compatibility with the pre-trained GOAL codebase, the data is stored in a compressed NumPy format (.npz) with the following specific arrays:

\begin{table}[t]
    \centering
    {
    \begin{tabular}{c | c | c | c | c | c  }
Problem &  Size & OR Solver & MTT & Optimality Gap  & time (s) \\
\hline
KP & & & &\\
         & 50 & 24040 & 23996 & 0.0019 & 16 \\
         & 60 & 27343 & 27292 & 0.0019 & 23 \\
         & 70 & 30074 & 30041 & {\bf 0.0011} & 33 \\
         & 80 & 32458 & 32411 & 0.0015 & 45 \\
         & 90 & 34723 & 34676 & 0.0014 & 59 \\
         & 100 & 36531 & 36490 & {\bf 0.0011} & 78 \\
\hline 
JSP & & & & & \\
& $5\times 5$  & 410 & 421 & 0.027 & 23 \\
& $6\times 6$  & 502 & 515 & {\bf 0.024} & 58\\
& $7\times 7$  & 580 & 594 & 0.025 & 131\\
& $8\times 8$  & 653 & 673 & 0.031 & 268\\
& $9\times 9$  & 726 & 752 & 0.035 & 516\\
& $10\times 10$  & 817 & 849 & 0.039 & 921\\

    \end{tabular}}
    
    \caption{Comparison of the solution quality gap between the combinatorial optimization solver ``OR solver", the multi-type Transformer solver ``MTT", the optimality gap of the MTT solver, and the inference time of the multi-type transformer on a single NVIDIA A100 GPU of 80GB memory. }
    \label{tab:bench}
\end{table}

\begin{itemize}
    \item weights (Shape: $128 \times 90$ weights are the "Cost" or "Size.” And the total weight Must be below the capacity limit. A matrix containing the weight of every item in every instance. Generated uniformly in [0, 1] and scaled by 1000 (e.g., $0.5 \to 500$).
    \item values (Shape: $128 \times 90$): Values are the "Profit" or "Reward." There is no limit on the value. The target is to make the total value as high as possible. A matrix containing the value of every item. Generated like weights.
    \item Capacities \(Shape: 12\): A vector containing the capacity constraint for each of the 128 instances Set to 20,000 (equivalent to 20.0 in normalized terms).
\item Optimal value (Shape: 128): The ground-truth maximum value achievable for each instance, calculated using the ORTools SCIP solver. This serves as the denominator for calculating the Optimality Gap.
\item Scale (scaler value: 1000): A normalization factor used to convert between integer representations and the normalized float values that the model expects. The input file uses large integers to avoid floating-point errors and compatibility issues in the Python data loader.
\end{itemize}

There is a noted discrepancy between the original paper text (stating a fixed capacity of 25) and the official repository's test files (using a scaled capacity of 20,000, effectively 20).
To align with the pre-trained model's actual training distribution—rather than the potentially outdated text—the instances using the Fixed Capacity of 20,000 were generated. This ensures that the "tightness" of the constraint matches what the model "learned" during training, providing a fair baseline for evaluating.

Similarly, the Job-Shop Scheduling problem is tested on configurations of $10\times 10$, $9\times 9$, $8\times 8$, $7\times 7$, $6\times 6$, and $5\times 5$ jobs and machines, capturing diverse scheduling scenarios with escalating combinatorial explosion. Addressing these benchmarks necessitates robust optimization strategies—ranging from exact solvers to heuristic and learning-based approaches—to ensure solution quality and efficiency across scales, see Table~\ref{tab:bench}.

The JSP is  the Taillard and Demirkol instances which are classic static JSP, plus dynamic arrivals synthesized with Poisson processes that provides mirror disjunctive graph representations. The KP involves synthetic 0-1 KP with correlated and uncorrelated value–weight distributions to reflect robustness; sizes scale from 10 to 100 items.
The results in Table~\ref{tab:bench} on only 128 instances shows MTT has relative loss gap in scale of .02 in JSP. This means the JSP is a more difficult problem the it gives a reltive loss almost $.02$ times compared with a combinatorial solver. However, for KP this relative gap is reduced to the range of $0.001$ times of combinatorial optimization solver, making it more suitable for the industrial use.   
\section{\uppercase{Ferro-Titanium Application}}
The application of MTT model is demonstrated through a real-world resource allocation problem in a Ferro-titanium manufacturer. This process involves charging an induction furnace with various grades of titanium scrap and raw materials to produce a specific alloy.
The primary objective is to minimize the total raw material cost for a single batch, subject to strict operational constraints. Unlike theoretical knapsack problems where capacity is an upper bound, the industrial process requires the furnace to be fully loaded to operate efficiently.
\subsection{Problem Definition}
The problem is defined by the following key parameters and constraints:
\begin{itemize}
    \item Target Capacity ($W$) The furnace must receive exactly 1,800 lb of material.
    \item Material Inventory: There are 14 distinct types of raw materials available (e.g., Prepared solid 6-4, Mixed turnings, Titanium solids).
    \item Constraints:
    \begin{itemize}
        \item Cost: Each material has a specific unit cost (\$/lb), ranging from \$0.40 to \$1.10.
        \item Availability: Each material has a maximum allowable usage limit (lb/batch) due to supply availability or chemical composition limits.
    \end{itemize}
\end{itemize}
The challenge is to select the optimal combination of materials that sums exactly to the furnace capacity while yielding the lowest possible financial cost.

\subsection{Mapping to KP}

To utilize the MTT model which is trained to solve the classic 0-1 Knapsack Problem—the continuous minimization problem of blending required transformation into a discrete maximization problem but in reality we are facing the fractional knapsack problem which the MTT does not handle. This section details the data transformation pipeline to feed the MTT model.

In the physical facility, materials exist in continuous quantities (piles). However, KP is formalized as binary discrete, requiring distinct ``items'' to be selected or rejected. To bridge this gap, we simulated the raw material inventory as a collection of discrete batches or containers.

\begin{itemize}
    \item For each problem instance, we generated datasets with item counts ranging from $N=50$ to $N=100$ to simulate varying inventory granularities. This setting also aligns with our benchmark experiments.
    \item Each item represents a container of a specific material type $k$.
    \item The weight of each item, $w_i$, was sampled uniformly from a realistic range (e.g., 20 lb to 200 lb), ensuring the total weight of items of type $k$ did not exceed the maximum usage limit defined in the industrial specifications.
\end{itemize}
\subsection{Objective Alignment}
Standard MTT models are designed to maximize total value. Our objective, however, is to minimize total cost. To align these objectives, we introduce a \textit{Reference Price Transformation}. Let's define the reference price as $P_\text{ref}$ (set at $\$2.0/\text{lb}$), which represents a theoretical maximum willingness-to-pay, strictly higher than the most expensive raw material ($\$1.10/\text{lb}$). The ``value'' $v_i$ for the MTT model was calculated as the ``savings'' generated by using a specific material compared to the reference price as follows
\begin{equation}
    v_i = w_i (P_\text{ref} - c_i)  ,
\end{equation}

\noindent where
\begin{itemize}
    \item $c_i$ is the actual unit cost of the material ($\$/\text{lb}$),
    \item $w_i$ is the weight of the item ($\text{lb}$).
\end{itemize}

Lower cost materials yield a higher $v_i$ (virtual profit). By maximizing $\sum v_i$, the MTT model implicitly minimizes $\sum c_i w_i$, effectively solving the cost minimization problem. The industrial process requires the total weight to equal the capacity ($\sum w_i = W$), whereas the standard KP enforces an inequality ($\sum w_i \leq W$).

By ensuring that all transformed values $v_i$ are strictly positive via $P_\text{ref} > \max(c_i)$, the model is naturally incentivized to fill the knapsack as much as possible to accumulate reward. This drives the total weight toward the capacity limit of 1,800 lb. Any minor underfilling in the model's output (e.g., reaching 1,790 lb) represents a small ``optimality gap'' or requires a trivial post-processing step (adding a small amount of filler material) in deployment.

\subsection {Numerical Results}
Ground truth values calculated using the OR-Tools solver in PyTorch to serve as a baseline for evaluating the MTT optimality gap.
\begin{table}[h]
    \centering
    {
    \begin{tabular}{c | c | c | c | c | c  }
Problem &  Size & OR Solver & MTT & Optimality Gap & time (s) \\
\hline
KP & & & &\\
         & 50 & 2728& 2647& 0.029&09 \\
         & 60 &2743 & 2670 &0.026 &12 \\
         & 70 &2753 & 2677 & 0.027& 15 \\
         & 80 & 2765& 2693 &0.026 & 18 \\
         & 90 & 2768& 2697 & {\bf 0.025}  &23 \\
         & 100 & 2779&2705 & 0.026& 28\\
    \end{tabular}}
\caption{Breakdown of the Ferro-titanium application mapped to KP with different equal item size ranging from 50 to 100 items, suggesting devide the load into 90 items for a better MTT solution quality.}
\label{tab:ferro}
\end{table}
Table~\ref{tab:ferro} summarizes the performance of MTT model across varying inventory densities ($N=50$ to $100$). The model demonstrates remarkable stability, maintaining an optimality gap between $0.025$ and $0.029$ relative change across all test cases. While the neural network approximates the solution significantly faster than exact solvers, it consistently captures roughly the optimal loss of the combinatorial optimization solver with the potential theoretical savings in parallel computation thanks to the Transformer architecture.

The peak performance was observed at $N=90$, where the model achieved its lowest gap of $0.025$. Inventory size of 90 batches provides the optimal balance of ``granularity'' for the model to find efficient packing combinations without being overwhelmed by the search space.


\section{\uppercase{Conclusion}}
This work demonstrates that Multi‑Type Transformers (MTT) offer a unified and scalable framework for tackling structurally diverse combinatorial optimization problems relevant to enterprise resource planning (ERP), specifically the 0-1 Knapsack Problem (KP) and the Job‑Shop Scheduling Problem (JSP). By formulating both problems within a heterogeneous graph representation, the model effectively leverages type‑specific attention to capture the distinct relational structures inherent to packing and scheduling tasks. Across our benchmarks, MTT achieved competitive solution quality, with optimality gaps on the order of 0.001 for KP and approximately $0.03$ for JSP, indicating strong performance on static packing problems and promising potential for more complex industrial planning. Beyond synthetic benchmarks, we extended the methodology to a real industrial application in the Ferro‑Titanium manufacturing process. Through a principled data‑generation pipeline and cost‑reformulation strategy, the MTT model produced near‑optimal material‑loading plans with stable gaps around $0.025-0.029 $, while requiring less than a second for each instance on a GPU. These findings affirm that multi‑type attention architectures can generalize beyond academic benchmarks and provide actionable decision support in real manufacturing environments. 

Our findings confirm that multi‑type attention architectures offer a promising direction for learning‑based ERP optimization. The study also reveals areas requiring deeper investigation. In particular, a more comprehensive evaluation including comparisons with classical heuristics, neural baselines, ablations of the architectural components, and a standardized reporting of optimality gaps and runtimes would strengthen the empirical claims. 

Overall, this study provides evidence that MTT can effectively address certain ERP‑motivated optimization tasks and can be adapted to industrial pipelines. At the same time, the feedback from reviewers highlights important avenues for improving rigour, transparency, and comparative evaluation in future iterations of this research.

\bibliographystyle{apalike}
{\small
\bibliography{ref}}



\end{document}